\documentclass[final]{cvpr}

\usepackage{times}
\usepackage{epsfig}
\usepackage{graphicx}
\usepackage{amsmath}
\usepackage{amssymb}
\usepackage{amsfonts}

\usepackage{mathabx}
\usepackage{array}
\usepackage{xcolor}
\newcolumntype{M}[1]{>{\centering\arraybackslash}m{#1}}
\newcolumntype{L}[1]{>{\flushleft\arraybackslash}m{#1}}
\definecolor{mygray}{gray}{0.0}

\usepackage[pagebackref=true,breaklinks=true,colorlinks,bookmarks=false]{hyperref}

\def\cvprPaperID{2227} 
\def\confYear{CVPR 2021}

\begin{document}

\title{Towards Unified Surgical Skill Assessment}

\author{
    Daochang Liu$^{1,3,5}$, 
    Qiyue Li $^1$, 
    Tingting Jiang$^1$, 
    Yizhou Wang$^{1,4}$, 
    Rulin Miao$^2$, 
    Fei Shan$^2$, 
    Ziyu Li$^2$
    \vspace{3pt}\\
    $^1$NELVT, Department of Computer Science, Peking University\\
    $^2$Peking University Cancer Hospital, $^3$Deepwise AI Lab\\
    $^4$Center on Frontiers of Computing Studies, Peking University\\
    $^5$Advanced Institute of Information Technology, Peking University\\
    {\tt\small \{daochang, liqiyue, ttjiang\}@pku.edu.cn}\\
}

\maketitle
\pagestyle{empty}
\thispagestyle{empty}

\begin{abstract}
Surgical skills have a great influence on surgical safety and patients' well-being.
Traditional assessment of surgical skills involves strenuous manual efforts, which lacks efficiency and repeatability.
Therefore, we attempt to automatically predict how well the surgery is performed using the surgical video.
In this paper, a unified multi-path framework for automatic surgical skill assessment is proposed, which takes care of multiple composing aspects of surgical skills, including surgical tool usage, intraoperative event pattern, and other skill proxies.
The dependency relationships among these different aspects are specially modeled by a path dependency module in the framework.
We conduct extensive experiments on the JIGSAWS dataset of simulated surgical tasks, and a new clinical dataset of real laparoscopic surgeries.
The proposed framework achieves promising results on both datasets, with the state-of-the-art on the simulated dataset advanced from 0.71 Spearman's correlation to 0.80.
It is also shown that combining multiple skill aspects yields better performance than relying on a single aspect.
Codes at \url{https://git.io/JG2OA}.
\end{abstract}

\section{Introduction}

Hundreds of millions of surgeries are performed worldwide annually~\cite{2015_Lancet_Weiser}.
The proficiency of the operating surgeon is a key factor affecting outcomes after surgery~\cite{2013_NEJM_Birkmeyer}.
To ensure patient safety and reduce clinical errors, surgical skill assessment has become an indispensable part of surgical training~\cite{1993_AJS_Reznick} and credentialing~\cite{2002_CPS_Wanzel}.

Conventional surgical skill assessment is undertaken manually by experts with direct observation~\cite{1993_AJS_Reznick} or structured rating protocols~\cite{OSATS}.
Such human assessment is slow and hardly reproducible.
Meanwhile, the prevalence of laparoscopic and robot-assisted surgeries nowadays brings a large volume of surgery videos captured by the built-in cameras in surgical devices, which lay the foundation for automatic learning-based approaches to provide efficient and repeatable skill assessment~\cite{2017_ARBE_Vedula}.
This paper works on automatic surgical skill assessment using surgical videos.

\begin{figure}[t]
\begin{center}
   \includegraphics[width=0.9\linewidth]{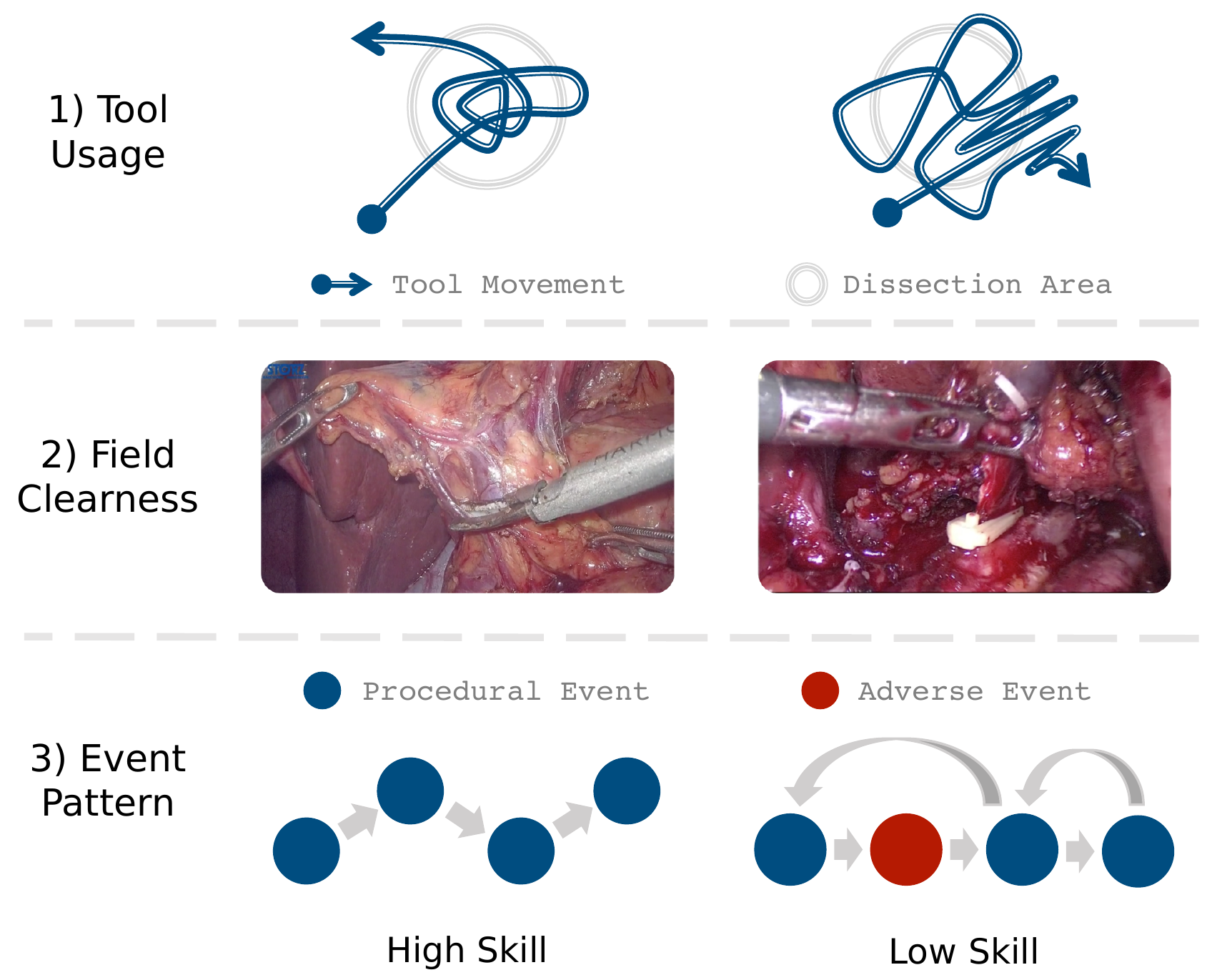}
\end{center}
   \caption{\textbf{Different aspects of surgical skills.} Surgical skills can be assessed from many aspects, \eg, 1) the usage of surgical tools 2) the clearness of the operating field 3) the distribution of surgical events. This paper proposes a unified framework for surgical skill assessment, which exploits these different aspects and the interaction among them. Best viewed in color.}
\label{fig:intro}
\end{figure}

Surgical skills are complex with many facets.
No universally accepted skill assessment criterion exists in the medical field currently~\cite{2019_JU_Chen}.
By discussion with clinicians, we identify three important aspects from the medical literature that are likely to characterize surgical skills and also suitable for automatic assessment, \ie, surgical tool usage, surgical field clearness, and surgical event pattern.
\textit{The first aspect} is the movement of surgical tools~\cite{2017_JSE_Ghasemloonia,2013_SE_Mason}, which could reflect the instrument handling proficiency and motion efficiency of the surgeon.
As in Fig.~\ref{fig:intro}, a high-skill surgeon will have a short and smooth tool trajectory concentrating on the dissection area, while a low-skill surgeon will have a lengthy and jerky trajectory dispersed in a large spatial range.
\textit{The second aspect} is the clearness of the operating field as a skill proxy~\cite{COF_MICCAI}.
Skill proxy means an indirect indicator that is statistically correlated to surgical skills.
Concretely, a clear operating field ensures high visibility of anatomy structures, which is critical for the surgeon's performance. 
And it is more likely to link a poorly-performed surgery to an obscure field with excessive bleeding, burnt tissue, smoke, and thus limited anatomy visibility.
\textit{The third aspect} is the workflow of surgical events or actions~\cite{2016_IJCARS_Uemura}.
For example in Fig.~\ref{fig:intro}, a well-performed surgery tends to have a linear pattern of events following the optimal procedure. 
On the contrary, a loopy pattern with more jumps across events is more common in a poorly-performed surgery, because the surgeon could find some previous surgical step unsatisfactory and go back to fix it. 
Besides, skill-related factors are also major causes of adverse events in surgeries, such as bleeding and injury~\cite{2011_PSS_Zegers}.

Prior works on automated surgical skill assessment, \eg~\cite{2009_MICCAI_Reiley,2018_WACV_Jin,COF_MICCAI}, mostly rely on one of these aspects.
However, we believe the great complexity of surgical skills necessitates a combination of multiple aspects for an accurate assessment.
Besides, the dependency relationships among different aspects also play important roles in skill assessment.
For instance, tool usage needs to be more careful when the field clearness is impaired.
This paper thus conceptualizes a unified framework to leverage the complementary information in different skill aspects and also capture dependencies among aspects.
The proposed framework comprises multiple paths in parallel, with each corresponding to a skill aspect.
Aspect-specific feature sequences extracted from surgical videos are forwarded along each path, subsequently transformed into skill score sequences for each aspect.
We integrate a path dependency module into the framework to capture inter-path dependencies.
In this module, feature sequences are aggregated from all the skill aspects to provide temporal importance weights for the score sequences.
Lastly, the weighted score sequences are pooled over time and fused across paths as the final assessment prediction.
One practical problem of surgical skill assessment is the limited amount of annotations for training.
Therefore, apart from a classic supervised regression loss, the framework is additionally equipped with a self-supervised contrastive loss to learn without annotations.
Specifically, we employ a predictive coding mechanism on the latent embedding of feature sequences.

On the other hand, existing approaches are usually validated on simulated surgical tasks, such as knot-tying in the JIGSAWS benchmark~\cite{JIGSAWS_1,JIGSAWS_2}.
However, a clinical dataset of real surgeries is more desirable.
And it is better to also have event and tool annotations to support multi-aspect assessment.
The clinical dataset in EndoVis Challenge 2019~\cite{EndoVis2019} satisfies these requirements, which is unfortunately not publicly usable yet.
Therefore, we collect a new clinical dataset consisting of twenty \textit{in vivo} laparoscopic gastrectomies with comprehensive skill and event annotations.
The proposed framework is validated by extensive experiments on our new clinical dataset and the simulated JIGSAWS dataset.
On both datasets, instantiations of the proposed framework obtain state-of-the-art results.
Experimental results are higher when multiple skill aspects are combined, validating the effectiveness of our unified approach.
We also correlate the predicted skill scores with the input features on the temporal dimension to get insights on how the model understands surgical skills. 

To summarize, our contributions are three-fold: 1) A unified framework assessing surgical skills from multiple aspects 2) A new clinical surgery dataset 3) Promising results on both simulated and clinical surgery datasets.


\section{Related Works}
\subsection{Automatic Surgical Skill Assessment}
Prior works roughly fall into three categories according to which skill aspect they are related to.
\textit{The first category} is tool-related and makes up the majority of the literature.
Methods in this category rely on tool motion data from various sources, including video object tracking or detection~\cite{2018_WACV_Jin,2019_AS_Azari,2020_IJCARS_Escamirosa}, video spatiotemporal descriptors~\cite{2018_IJCARS_Zia_2,2015_MICCAI_Zia,2014_PAMI_Zhang,2013_CVPR_Bettadapura}, robotic kinematics~\cite{2018_IJCARS_Zia,2018_IJCARS_Wang,2018_MICCAI_Fawaz,2019_IJCNN_Castro}, external sensors~\cite{2016_MICCAI_Ershad,2015_IJCARS_Ahmidi,2019_IJCARS_Holden}, and virtual reality interfaces~\cite{2011_ICRA_Jog}.
\textit{The second category} is proxy-related.
The clearness of operating field is identified as a skill proxy on clinical data~\cite{COF_MICCAI}.
In their method~\cite{COF_MICCAI}, indirect assessment via proxy performs better than direct skill prediction.
\textit{The third category} is event-related.
They usually break down surgical trials into fine-grained events~\cite{2009_MICCAI_Reiley,2012_IPCAI_Tao,2014_IPCAI_Malpani,2013_MICCAI_Ahmidi,2009_MICCAI_Varadarajan,2020_Surgery_Luongo}. 
These methods use surgical events differently from our framework. 
Our framework learns from the occurrence pattern of events to determine skills, while they mainly evaluate surgical skills in the individual event to provide detailed feedback.
A recent method~\cite{2020_MICCAI_Wang}, which also belongs to the event-related category, tackles surgical skill assessment by multi-task learning with surgical gestures.
In addition, some researchers adopt a purely learning-based approach~\cite{2019_IJCARS_Funke} that is not related to any of these three categories.

Most previous studies take advantage of only one skill aspect.
Our framework, instead, tries to unify multiple aspects for surgical skill assessment.

\begin{figure*}[t]
\begin{center}
   \includegraphics[width=1.0\linewidth]{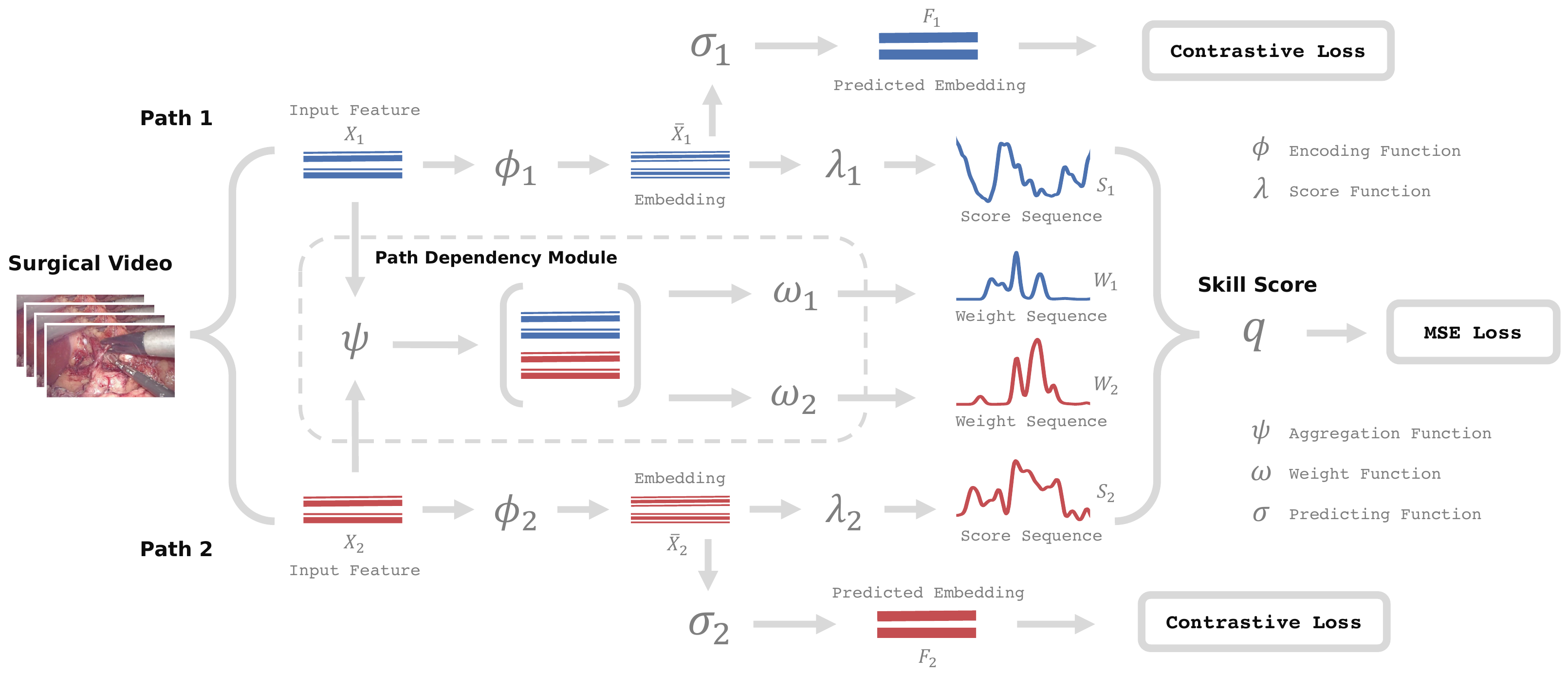}
\end{center}
   \caption{Multi-path framework for unified surgical skill assessment. Two paths are visualized for clarity and four paths are used in practice.}
\label{fig:overview}
\end{figure*}

\subsection{Action Quality Assessment}
Action quality assessment is a field relevant to surgical skill assessment.
Methods in this field aim at assessing the quality of actions in sports such as diving and gymnastics~\cite{2017_CVPR_Bertasius,2019_WACV_Parmar,2020_ECCV_Gao,2020_CVPR_Tang,2019_ICCV_Pan,2020_CVPRW_Nekoui,2020_MM_Zeng,2014_ECCV_Pirsiavash,2017_CVPRW_Parmar,2019_CVPR_Parmar,2018_ACCV_Li,1995_AIED_Gordon,2003_ICCVS_Jug,2019_TCSVC_Xu}, or actions in daily life such as drawing and going upstairs~\cite{2019_ICCVW_Li,2014_BMVC_Paiement,2019_CVPR_Doughty,2018_CVPR_Doughty}.
The quality assessment of simulated surgical tasks is involved in the experiments of several methods above~\cite{2018_CVPR_Doughty,2020_ECCV_Gao,2020_CVPR_Tang,2019_ICCV_Pan,2019_CVPR_Doughty}, but not as their central focuses.
More importantly, medical domain knowledge is not sufficiently incorporated in these general-purpose methods.
However, such domain knowledge is crucial for surgical skill assessment.

%

\section{Method}
In this section, we present a unified framework for surgical skill assessment, which takes in a surgery video and outputs a skill score.
As shown in Fig.~\ref{fig:overview}, our framework comprises multiple assessment paths, an inter-path dependency module, and a contrastive learning mechanism.
This section introduces the framework in its general form and detailed instantiations are left to the experiment section~\ref{sec:instantiation}. 

\subsection{Multi-Path Assessment}

To characterize surgical skills from multiple aspects, our framework adopts a multi-path design, in which multiple paths with similar architectures are organized in parallel such that each path concentrates on one skill aspect.
Concretely, four paths are included, with three of them corresponding to the previously mentioned skill aspects, \ie, tool, proxy, and event.
The rest one is a baseline visual path that evaluates surgical skills directly from semantic visual features, such as the features from pre-trained deep neural networks.
The input to each path is a feature sequence extracted from the surgical video, which is intended to supply distinct information specific to each skill aspect.
Feature sequences for different paths are of similar shapes:
\begin{equation}
X_{m} \in \mathbb{R}^{L \times D_{m}}, m \in \{\mathtt{V}, \mathtt{T}, \mathtt{P}, \mathtt{E}\}
\end{equation}
where $\mathtt{V}, \mathtt{T}, \mathtt{P}, \mathtt{E}$ denote the visual, tool, proxy, event path respectively, $X_{m}$ denotes the feature sequence input to the path $m$, $L$ is the video length and $D_{m}$ is the feature dimension.
The extraction of these features is flexible and can adapt to the dataset, task, application and so on, as long as each focuses on its skill aspect.
For example, the input to the event path can be a sequence of occurrence probabilities of surgical events, and the input to the tool path can be a sequence of spatial coordinates of surgical tools.

Along each path $m$, the feature sequence is first encoded into a high-level embedding sequence:
\begin{equation} \label{eq:2}
\bar{X}_{m} = \phi_{m}(X_{m})
\end{equation}
where $\phi_{m}$ represents an encoding function in the path $m$, and $\bar{X}_{m} \in \mathbb{R}^{L \times \bar{D}_{m}}$ is the resultant embedding with size $\bar{D}_{m}$.
Afterward, the embedding sequence is converted into a score sequence $S_{m} \in \mathbb{R}^{L \times 1}$ indicating predicted surgical skill at each time step:
\begin{equation}
S_{m} = \lambda_{m}(\bar{X}_{m}).
\end{equation}
The $\lambda_{m}$ denotes a score function in the path $m$. 
In this way, each path gives an aspect-specific rating of surgical skills.

\subsection{Path Dependency Module}
In the assessment of surgical skills, the relation among skill aspects matters.
Skill predictions in one path contribute to the overall assessment unequally at different time steps, often depending on the information in other paths.
For example, the surgical tool usage could become more important for skill assessment when the field clearness is reduced, and the event occurrence could become less important when no tool appears in the scene.
To model such inter-path dependencies, we design a path dependency module in our framework, which mimics the phenomena in these examples by gathering information from all the paths to provide different temporal importance weights for each path.
In detail, the importance weight $W_{m} \in \mathbb{R}^{L \times 1}$ for the path $m$ is also a temporal sequence.
To compute the weight $W_{m}$, feature sequences are collected from all the paths by an aggregation function $\psi$ and then sent to a weight function $\omega_{m}$:
\begin{equation}
W_{m} = \mathrm{softmax}(\omega_{m}(\psi(X_{\mathtt{V}}, X_{\mathtt{T}}, X_{\mathtt{P}}, X_{\mathtt{E}}))).
\end{equation}
The aggregation function is shared by paths while the weight functions are not.
And a softmax function is imposed to normalize the weight sequence on the temporal dimension.
Subsequently, the score sequences, weighted by the temporal importance, are averaged over time and paths to obtain an overall video-level skill score:
\begin{equation}
q = \frac{1}{4}\sum_{m} \sum_{i=0}^{L}{S_{m,i} W_{m,i}}, m \in \{\mathtt{V}, \mathtt{T}, \mathtt{P}, \mathtt{E}\}
\end{equation}
where $S_{m,i} \in \mathbb{R}$, $W_{m,i} \in \mathbb{R}$ denote the score and weight in the path $m$ at time $i$, and $q \in \mathbb{R}$ is the video-level skill prediction.
Lastly, a mean squared error (MSE) loss is utilized to supervise the skill prediction:
\begin{equation}
\mathcal{L}_{mse} =  {(y - q)}^{2}.
\end{equation}
The $y$ is the ground truth skill annotation, which is within a predefined range of scores.

\subsection{Self-Supervised Contrastive Loss}

The scarcity of annotated data is a common concern for medical tasks, and surgical skill assessment is no exception.
To alleviate this issue, we resort to a contrastive learning strategy.
Inspired by the video predictive coding~\cite{2020_ECCV_Han}, we take future prediction as an auxiliary task to help the model learn temporal dynamics in a self-supervised manner.
Concretely, in each path $m$, a predicting function $\sigma_{m}$ is used to forecast the embedding in the future based on the recent past:
\begin{equation}
F_{m,i} = \sigma_{m}(\bar{X}_{m,i-1})
\end{equation}
where $\bar{X}_{m,i-1} \in \mathbb{R}^{\bar{D}_{m}}$ is the embedding in Eqn.~\ref{eq:2} at time $i-1$, and $F_{m,i} \in \mathbb{R}^{\bar{D}_{m}}$ is the predicted embedding for time $i$.
Then a contrastive loss is designed to encourage the similarity of the predicted embedding $F_{m,i}$ with the real embedding $\bar{X}_{m,i}$ at time $i$, and discourage its similarity with the embeddings in other time steps:
\begin{equation}
\mathcal{L}_{con} = - \sum_{m} \sum_{i=1}^{L} \mathrm{log} \frac{\mathrm{exp}(F_{m,i} \cdot \bar{X}_{m,i})} {\sum_{j \in \mathcal{N}_{i}} \mathrm{exp}(F_{m,i} \cdot \bar{X}_{m,j})}
\end{equation}
where $\cdot$ represents the dot product, and $\mathcal{N}_{i}$ is a temporal neighborhood around time $i$ including time $i$.
This contrastive loss could assist the encoding function $\phi_{m}$ in Eqn.~\ref{eq:2} in better capturing temporal dynamics in the surgical video.

Finally, we combine the self-supervised contrastive loss with the supervised MSE loss to train the framework:
\begin{equation}
\mathcal{L}_{full} =  \mathcal{L}_{mse} + \mathcal{L}_{con}.
\end{equation}


\section{Experiments}

This section first introduces the experimental setup and the implementation of our framework.
We then present ablation studies and comparisons to state-of-the-art methods.
At last, the correlation between model prediction and input features is examined. 

\begin{figure}[b]
\begin{center}
   \includegraphics[width=1\linewidth]{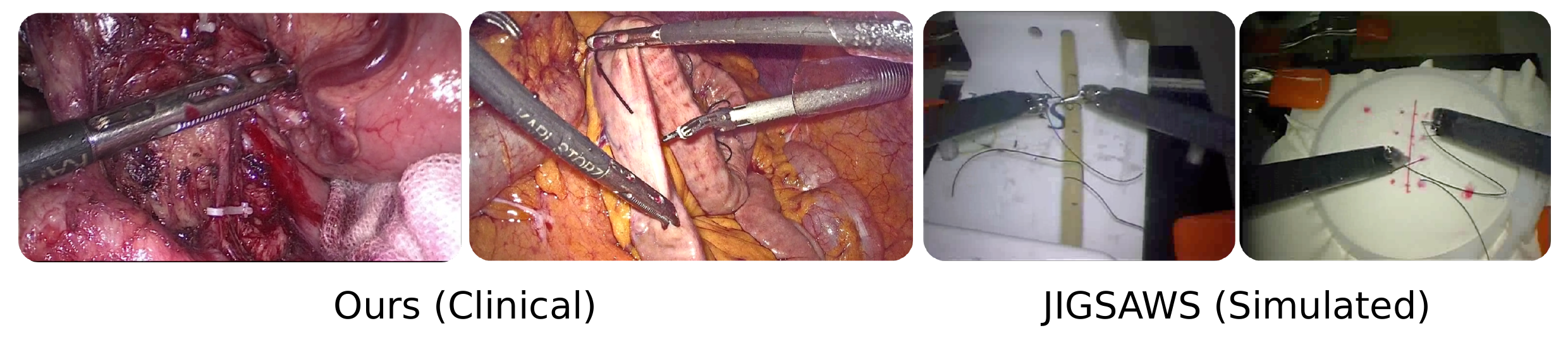}
\end{center}
   \caption{Example video frames.}
\label{fig:example}
\end{figure}

\subsection{Datasets}

\textbf{Simulated dataset.} 
Experiments are first performed on the public JIGSAWS dataset~\cite{JIGSAWS_1,JIGSAWS_2}, which contains three simulated tasks for robotic-assisted surgery, \ie, suturing (SU), needle-passing (NP), and knot-tying (KT).
There are 78 egocentric videos for the suturing task, 56 for needle passing, and 72 for knot tying, with 206 videos in total.
The duration of the video is 88 seconds on average.
Each video is annotated with a skill-level global rating score (GRS) with a range from 6 to 30.
We use the GRS as the ground truth of the surgical skill.
The JIGSAWS dataset also provides annotations of fine-grained surgical gestures and kinematic data of robotic manipulators.

\textbf{Clinical dataset.}
Experiments are also performed on a newly built clinical dataset.
This dataset has 20 laparoscopic videos of \textit{in vivo} surgeries for gastric cancer, including partial or total gastrectomy and related lymph node (LN) dissection.
The videos have $960 \times 540$ resolution and 25 FPS.
Different from the simulated JIGSAWS dataset, our dataset is collected from real operating rooms.
This new dataset is very challenging, due to its extremely long duration (199 min. per video on average), frequent camera movement, changing illumination, and varying patient conditions.
Example frames are given in Fig.~\ref{fig:example}.

\begin{figure}[t]
\begin{center}
   \includegraphics[width=0.7\linewidth]{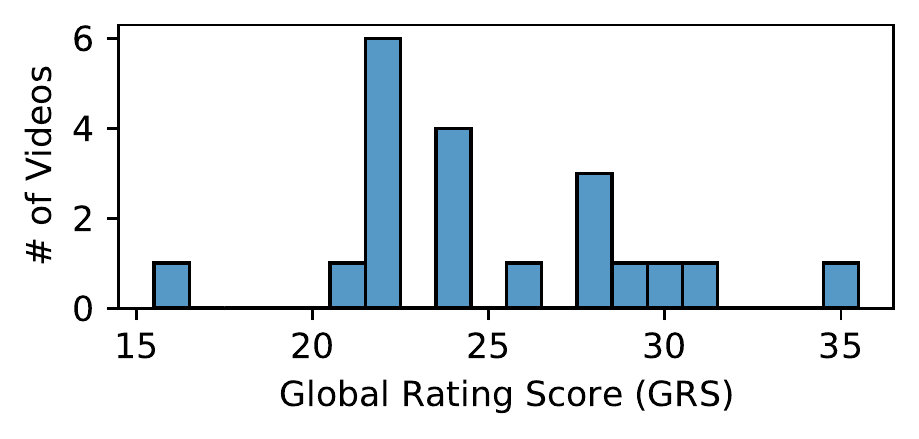}
\end{center}
   \caption{Skill distribution of the new clinical dataset.}
\label{fig:GRS_dist}
\end{figure}

For each video in our dataset, surgical skills are annotated by an expert surgeon on 7 metrics based on a modified OSATS protocol~\cite{OSATS}.
The global rating score (GRS) is defined as the sum of the 7 metrics.
The GRS has a range from 7 to 35 and is used as the ground truth.
In Fig.~\ref{fig:GRS_dist}, we plot the distribution of the GRS over the 20 videos. 
This dataset is also annotated with the proxy score of field clearness~\cite{COF_MICCAI}, as well as comprehensive surgical events.
There are 41 classes of surgical events in total, including procedural events, adverse events, video events, \etc.
The statistics of our dataset are summarized in Table~\ref{table:dataset} and more details are given in the supplementary material.
We use the 15 event classes listed in Table~\ref{table:event} in this study.

\begin{table}[t]
\begin{center}
\small
\begin{tabular}{M{0.55\linewidth} | M{0.15\linewidth}}
\hline
\# Videos & 20 \\
\# Average frames per video & 299K \\
Average duration per video & 199 min. \\
\hline
\# Surgical event classes & 41 \\
\# Surgical event instances & 1565 \\
\hline
\# Skill metrics & 7 \\
\# Skill proxy & 1 \\
\hline
\end{tabular}
\end{center}
\caption{Statistics of the new clinical dataset.} 
\label{table:dataset}
\end{table}

\begin{table}[t]
\begin{center}
\small
\begin{tabular}{M{0.86\linewidth} | M{0.05\linewidth}}
\hline
Surgical Event & \#Ins. \\
\hline
Abdominal cavity exploration & 31 \\
Dissection of fusion tissue & 19 \\
Dissection of the greater omentum & 24 \\
LN dissection of subpyloric region (SR) & 22 \\
LN dissection of hepatoduodenal ligament region (HLR) & 41 \\
LN dissection of the superior pancreas (SP) & 27 \\
LN dissection of lesser curvature (LC) & 21 \\
LN dissection of the left gastroepiploic region (LGR) & 22 \\
Resection of the distal stomach & 20 \\
Specimen removal & 20 \\
Gastro-jejunal anastomosis & 21 \\
Jejuno-jejunal anastomosis & 21 \\
Irrigation and placement of the drains & 17 \\
Bleeding & 279 \\
Camera out & 352 \\
\hline
\end{tabular}
\end{center}
\caption{Clinical surgical events used in this study.} 
\label{table:event}
\end{table}

\subsection{Experiment Setup} \label{sec:setup}
To keep consistent with existing literature, we adopt both four-fold cross-validation (4-Fold) and leave-one-user-out cross-validation (LOUO) when evaluating on JIGSAWS.
The four-fold splits provided by~\cite{2020_CVPR_Tang} and the LOUO splits associated with the JIGSAWS dataset are used.
On our clinical dataset, three-fold cross-validation is adopted.

Following prior works, we choose Spearman’s rank correlation (SROCC) as the evaluation metric.
For the JIGSAWS dataset, the average correlation across the three surgical tasks is computed by Fisher’s z-value~\cite{2019_WACV_Parmar}.
Besides, experiment results are averaged over multiple runs.

\subsection{Framework Instantiation} \label{sec:instantiation}
\subsubsection{Instantiation of input features}
Input features to the paths in our framework are instantiated differently to carry aspect-specific information as follows.

\textbf{Visual path input $X_{\mathtt{V}}$.}
We leverage the semantic features extracted from ResNet-101~\cite{ResNet} pre-trained on ImageNet as $X_{\mathtt{V}}$.
The feature dimension $D_{\mathtt{V}}$ is 2048.

\textbf{Tool path input $X_{\mathtt{T}}$.}
On the clinical dataset, to capture the surgical tool movement, we first apply an unsupervised tool segmentation method~\cite{2020_MICCAI_Liu}.
Then $X_{\mathtt{T}}$ is spatial histograms of the segmentation masks.
Specifically, the mask in each frame is divided into $3 \times 3$, $4 \times 4$, and $5 \times 5$ spatial grids.
The percentage of pixels belonging to surgical tools in each grid cell is taken as the feature.
The feature dimension $D_{\mathtt{T}}$ thus equals $9+16+25 = 50$.
Since the segmentation method is unsupervised, $X_{\mathtt{T}}$ does not involve extra data or annotations.
On the simulated dataset, the robotic kinematic data associated with the dataset is used as $X_{\mathtt{T}}$.
We use $D_{\mathtt{T}}=14$ dimensions, including the 3D positions, 3D velocities, and gripper angles of the two patient-side manipulators.

\textbf{Proxy path input $X_{\mathtt{P}}$.}
On the clinical dataset, the field clearness is used as a skill proxy.
The frame-level scores of the proxy are extracted from a re-implementation of~\cite{COF_MICCAI} as $X_{\mathtt{P}}$.
The training and testing of method~\cite{COF_MICCAI} follow the same cross-validation settings as in Section~\ref{sec:setup}.
Since the field clearness only works for clinical data, we employ another simple skill proxy for the simulated dataset, \ie, task completion time~\cite{2018_CVPR_Doughty}.
The $X_{\mathtt{P}}$ is set as a sequence with a constant value inversely proportional to the video length.
On the simulated dataset, $X_{\mathtt{P}}$ does not involve extra data or annotations.
$D_{\mathtt{P}}$ is 1 on both datasets.

\textbf{Event path input $X_{\mathtt{E}}$.}
For the event path, we train Multi-Stage Temporal Convolutional Networks (MS-TCN)~\cite{2019_CVPR_Farha} to detect surgical events on the clinical dataset or surgical gestures on the simulated dataset.
The event detection models are trained under the same cross-validation settings as in Section~\ref{sec:setup}.
The $X_{\mathtt{E}}$ is then set as the frame-level probabilities of events or gestures.
Its dimension $D_{\mathtt{E}}$, which equals the number of classes, is 10 for suturing, 8 for needle-passing, 6 for knot-tying on the simulated dataset, or 15 on the clinical dataset.

\subsubsection{Instantiation of functions}

The aggregation function $\psi$ is instantiated with the concatenation of the feature dimension.
For all paths except the proxy path, the encoding functions $\phi$ are instantiated with Temporal Convolutional Networks (TCN)~\cite{2017_CVPR_Lea}.
The score functions $\lambda$, weight functions $\omega$, predicting functions $\sigma$ are chosen as frame-wise multilayer perceptrons (MLP).
For the proxy path, since the input feature already represents the skill, the encoding function $\phi_{\mathtt{P}}$ and the score function $\lambda_{\mathtt{P}}$ are set as identity functions, and the weight function $\omega_{\mathtt{P}}$ is a constant function giving uniform weights at all time steps.
The contrastive loss and the predicting function are removed from the proxy path.

\subsubsection{Other implementation details}
The proposed framework is implemented using PyTorch~\cite{PyTorch}.
Model parameters are trained using mini-batch stochastic gradient descent with the Adam optimizer~\cite{Adam}.
The embedding sizes $\bar{D}_{\mathtt{V}}, \bar{D}_{\mathtt{T}}, \bar{D}_{\mathtt{P}}, \bar{D}_{\mathtt{E}}$ are set to $20, 4, 1, 4$ respectively. 
We freeze the input features after extraction.
This allows the model to have a larger temporal receptive field covering more video frames, which is necessary when handling the extremely long clinical videos and when learning long-term event patterns.
The simulated videos are sampled at 5 FPS and the clinical videos are sampled at 0.5 FPS.
The GRS is normalized within 0 and 1 during training.
Our codes will be released to offer other details. 


\begin{table}[t]
\begin{center}
\small
\begin{tabular}{M{0.20\linewidth} | M{0.12\linewidth} | M{0.08\linewidth} M{0.10\linewidth} M{0.08\linewidth} M{0.08\linewidth}} 
\hline
\hline
  & Clinical & \multicolumn{4}{c}{Simulated 4-Fold} \\
Method $\downarrow$ & 3-Fold & SU & NP & KT & Avg. \\
\hline
Ours ($\mathtt{V}$) & 0.201 & 0.642 & 0.666 & 0.729 & 0.681 \\
Ours ($\mathtt{T}$) & 0.250 & 0.765 & 0.566 & 0.662 & 0.673 \\
Ours ($\mathtt{P}$) & 0.469 & 0.396 & 0.333 & 0.803 & 0.554 \\
Ours ($\mathtt{E}$) & 0.241 & 0.603 & 0.200 & 0.762 & 0.560 \\
Ours ($\mathtt{VT}$) & 0.268 & 0.735 & 0.737 & 0.706 & 0.726 \\
Ours ($\mathtt{VTP}$) & 0.525 & 0.791 & \textbf{0.761} & 0.784 & 0.779 \\
Ours ($\mathtt{VTPE}$) & \textbf{0.565} & \textbf{0.834} & 0.756 & \textbf{0.819} & \textbf{0.805} \\
\hline
\hline
\end{tabular}
\end{center}
\caption{Ablation study on framework paths.} 
\label{table:ablation_study_path}
\end{table}

\begin{table}[t]
\begin{center}
\small
\begin{tabular}{M{0.24\linewidth} | M{0.12\linewidth} | M{0.08\linewidth} M{0.08\linewidth} M{0.08\linewidth} M{0.08\linewidth}}
\hline
 & Clinical & \multicolumn{4}{c}{Simulated 4-Fold} \\
 Method $\downarrow$ & 3-Fold & SU & NP & KT & Avg. \\
\hline
Full Model & \textbf{0.565} & 0.834 & \textbf{0.756} & \textbf{0.819} & \textbf{0.805} \\
Without $\omega$ & 0.509 & 0.808 & 0.681 & 0.754 & 0.752 \\
Without $\psi$ & 0.522 & 0.766 & 0.555 & 0.777 & 0.712 \\
Without $\sigma$ & 0.414 & \textbf{0.853} & 0.676 & 0.797 & 0.786 \\
\hline
\end{tabular}
\end{center}
\caption{Ablation study on framework components.} 
\label{table:ablation_study_component}
\end{table}

\begin{table}[t]
\begin{center}
\small
\begin{tabular}{M{0.30\linewidth} | M{0.10\linewidth} M{0.10\linewidth} M{0.10\linewidth} M{0.10\linewidth}} 
\hline
Clinical 3-Fold  & \multicolumn{4}{c}{Simulated 4-Fold (Acc. \%)} \\
(mAP \%) & SU & NP & KT & Avg. \\
\hline
75.4 & 88.8 & 78.3 & 85.4 & 84.2 \\
\hline
\end{tabular}
\end{center}
\caption{Mean average precision (mAP) of surgical event detection and accuracy of surgical gesture detection.} 
\label{table:result_event}
\end{table}

\subsection{Ablation Studies}

\textbf{Effects of framework paths.}
A set of comparative experiments are performed on both datasets to inspect the effect of each skill aspect.
The results of using every single path and combinations of multiple paths are reported in Table~\ref{table:ablation_study_path}.
When using a single path, the proxy path achieves higher results than other paths on the clinical data.
In general, combining multiple paths improves performance.
On both two datasets, the best average performance is obtained when all the paths are included.
It is noticed that the event path performs badly on the needle-passing task, probably due to the inferior gesture detection accuracy on this task.
The accuracy of gesture detection and the mean average precision (mAP) of event detection are reported in Table~\ref{table:result_event}.

\begin{table}[t]
\begin{center}
\small
\begin{tabular}{M{0.60\linewidth} | M{0.20\linewidth}}
\hline\hline
Method & SROCC \\
\hline
USDL~\cite{2020_CVPR_Tang} & 0.161 \\
Ours ($\mathtt{VT}$) & \textbf{0.268} \\
\hline
MICCAI 2019~\cite{COF_MICCAI} & 0.469 \\ 
Ours ($\mathtt{VTP}$) & \textbf{0.525} \\
\hline
Ours ($\mathtt{VTPE}$) & \textbf{0.565} \\
\hline\hline
\end{tabular}
\end{center}
\caption{Comparison to the state-of-the-art on our clinical dataset. Methods in the same vertical slot can be directly compared.} 
\label{table:result_clinical}
\end{table}

\begin{table}[t]
\begin{center}
\small
\begin{tabular}{M{0.26\linewidth} | M{0.08\linewidth} | M{0.08\linewidth} M{0.08\linewidth} M{0.08\linewidth} M{0.08\linewidth}}
\hline\hline
Method & Input & SU & NP & KT & Avg. \\
\hline
USDL~\cite{2020_CVPR_Tang} & $\mathbb{V}$ & 0.64 & 0.63 & 0.61 & 0.63 \\
Ours ($\mathtt{VP}$) & $\mathbb{V}$ & \textbf{0.68} & \textbf{0.71} & \textbf{0.80} & \textbf{0.73} \\
\hline
{\color{mygray} MUSDL~\cite{2020_CVPR_Tang}} $\ast$ & {\color{mygray}$\mathbb{V}$} & {\color{mygray} 0.71} & {\color{mygray} 0.69} & {\color{mygray} 0.71} & {\color{mygray} 0.70} \\
\hline
ST-GCN~\cite{2019_ICCV_Pan,2018_AAAI_Yan} & $\mathbb{VK}$ & 0.31 & 0.39 & 0.58 & 0.43 \\
TSN~\cite{2019_ICCV_Pan,2018_CVPR_Doughty,TSN} & $\mathbb{VK}$ & 0.34 & 0.23 & 0.72 & 0.46 \\
JRG~\cite{2019_ICCV_Pan} & $\mathbb{VK}$ & 0.36 & 0.54 & 0.75 & 0.57 \\
AIM~\cite{2020_ECCV_Gao} & $\mathbb{VK}$ & 0.63 & 0.65 & \textbf{0.82} & 0.71 \\
Ours ($\mathtt{VTP}$) & $\mathbb{VK}$ & \textbf{0.79} & \textbf{0.76} & 0.78 & \textbf{0.78} \\
\hline
Ours ($\mathtt{VTPE}$) & $\mathbb{VK}$ & \textbf{0.83} & \textbf{0.76} & \textbf{0.82} & \textbf{0.80} \\
\hline\hline
\end{tabular}
\end{center}
\caption{Comparison to the state-of-the-art methods on the simulated dataset under the 4-Fold setting. Methods in the same vertical slot can be directly compared. $\mathbb{V}$: Surgical videos. $\mathbb{K}$: Robotic kinematics. $\ast$: Extra fine-grained skill annotations are used.} 
\label{table:result_simulated_4fold}
\end{table}

\begin{table}[t]
\begin{center}
\small
\begin{tabular}{M{0.39\linewidth} | M{0.07\linewidth} | M{0.06\linewidth} M{0.06\linewidth} M{0.06\linewidth} M{0.06\linewidth}}
\hline\hline
Method & Input & SU & NP & KT & Avg. \\
\hline
DTC+DFT+ApEn~\cite{2018_IJCARS_Zia} & $\mathbb{K}$ & 0.37 & 0.25 & \textbf{0.60} & 0.41 \\
Ours ($\mathtt{TP}$) & $\mathbb{K}$ & \textbf{0.40} & \textbf{0.63} & 0.55 & \textbf{0.53} \\
\hline
JRG~\cite{2019_ICCV_Pan} & $\mathbb{VK}$ & 0.35 & \textbf{0.67} & 0.19 & 0.40 \\
AIM~\cite{2020_ECCV_Gao} & $\mathbb{VK}$ & \textbf{0.45} & 0.34 & \textbf{0.61} & 0.47 \\
Ours ($\mathtt{VTP}$) & $\mathbb{VK}$ & \textbf{0.45} & 0.62 & 0.58 & \textbf{0.56} \\
\hline
{\color{mygray} MTL-VF (ResNet)~\cite{2020_MICCAI_Wang}} $\ast$ & {\color{mygray}$\mathbb{V}$} & {\color{mygray} 0.68} & {\color{mygray} 0.48} & {\color{mygray} 0.72} & {\color{mygray} 0.64} \\
{\color{mygray} MTL-VF (C3D)~\cite{2020_MICCAI_Wang}} $\ast$ & {\color{mygray}$\mathbb{V}$} & {\color{mygray} 0.69} & {\color{mygray} 0.86} & {\color{mygray} 0.83} & {\color{mygray} 0.80} \\
\hline
Ours ($\mathtt{VTPE}$) & $\mathbb{VK}$ & \textbf{0.45} & 0.65 & 0.59 & \textbf{0.57} \\
\hline\hline
\end{tabular}
\end{center}
\caption{Comparison to the state-of-the-art methods on the simulated dataset under the LOUO setting. Methods in the same vertical slot can be directly compared. $\mathbb{V}$: Surgical videos. $\mathbb{K}$: Robotic kinematics. $\ast$: Extra surgical experience annotations are used.} 
\label{table:result_simulated_louo}
\end{table}

\begin{figure*}[t]
\begin{center}
   \includegraphics[width=0.97\linewidth]{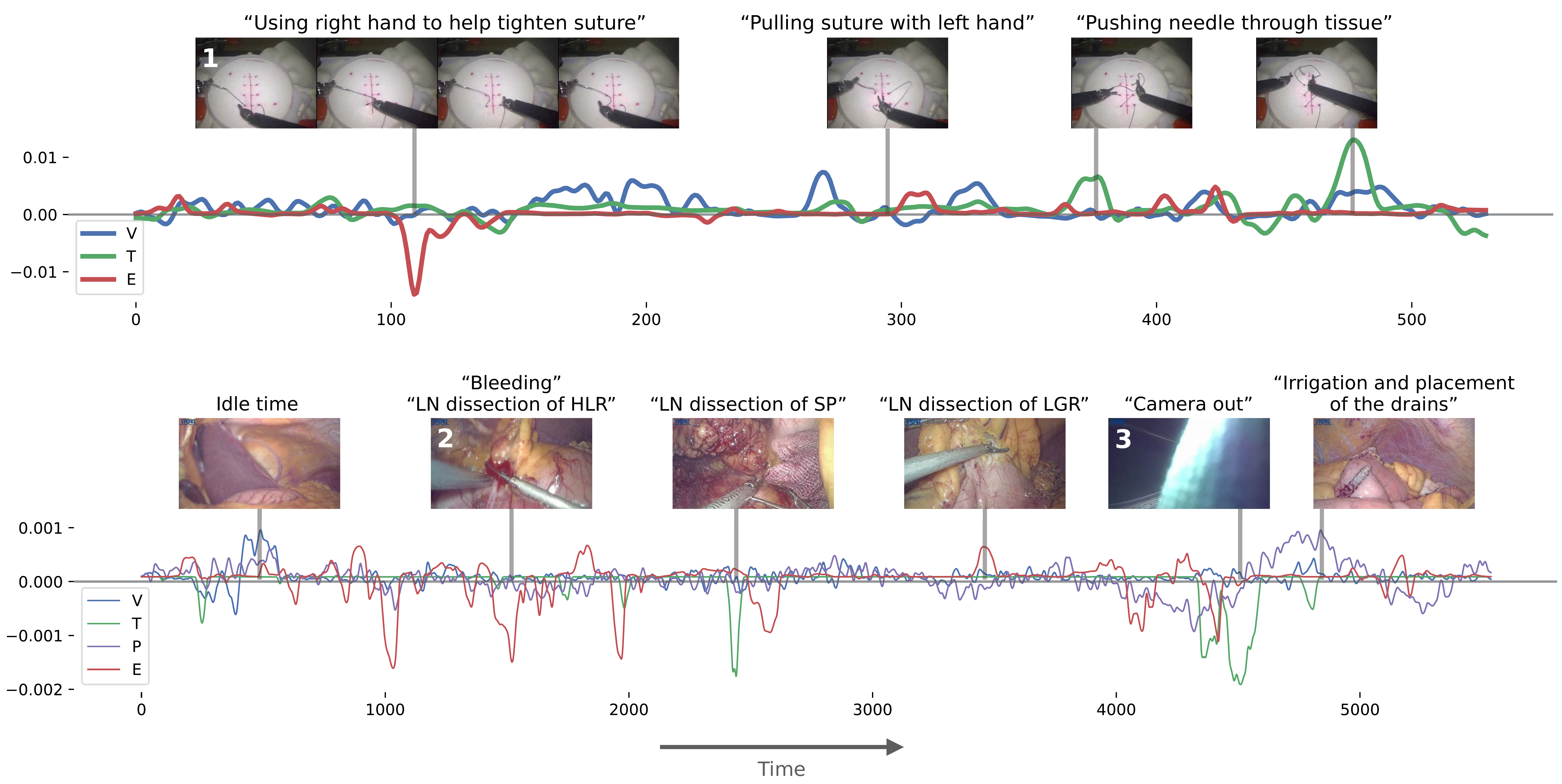}
\end{center}
   \caption{Result visualization. The upper part shows the result on a simulated surgery and the lower part on a clinical surgery. The weighted score sequences $S_{m}W_{m}$ from each path $m \in \{\mathtt{V},\mathtt{T},\mathtt{P},\mathtt{E}\}$ are visualized. The higher score reflects the better surgical skill and vice versa. Corresponding surgical gestures or events are marked on the selected frames. $S_{\mathtt{P}}W_{\mathtt{P}}$ is not plotted for the simulated surgery because it is a constant sequence. Note that for each sequence the integral over time equals a video-level skill prediction.}
\label{fig:visualization}
\end{figure*}

\textbf{Effects of framework components.}
In Table~\ref{table:ablation_study_component}, the results of our framework with one of the following components removed are presented: 1) weight functions $\omega$, 2) aggregation function $\psi$, 3) predicting function $\sigma$.
The contrastive loss is also discarded when the $\sigma$ is removed.
In general, all components contribute positively to the best performance.
Besides, the predicting function $\sigma$ for contrastive learning is especially important when the data is highly scarce, \ie, on the clinical data.
On the simulated suturing task, the contrastive learning yields no improvement possibly because sufficient training data is available.
The encoding and score functions are not ablated over since they lie in the backbone of our framework.


\subsection{Comparisons to State-of-the-Art}

\textbf{Clinical dataset.}
Due to the lack of existing results on our clinical dataset, two state-of-the-art methods are implemented to compare with, \ie, USDL~\cite{2020_CVPR_Tang} and MICCAI 2019~\cite{COF_MICCAI}. 
Specifically, we run the public code of USDL and re-implement the method~\cite{COF_MICCAI}. 
Moreover, for fairness, paths involving extra annotations are removed from our framework respectively in each comparison. 
We remove the proxy and event paths while comparing with USDL, and remove the event path while comparing with MICCAI 2019. 
Our method achieves promising performance as shown in Table~\ref{table:result_clinical}.

\textbf{Simulated dataset.}
Table~\ref{table:result_simulated_4fold} and Table~\ref{table:result_simulated_louo} show the comparisons between the experimental results of our method and other approaches on the simulated dataset. 
Note that previous methods could adopt different modalities as input, some of which use extra annotations. 
Therefore, methods using the same data and annotation are grouped together for comparison. 
When comparing our framework with others, we remove the paths involving extra annotations.
In each comparison, our method outperforms other counterparts respectively.

\subsection{Visualization}

We choose two videos from the simulated dataset and our clinical dataset to visualize the weighted score sequences $S_{m}W_{m} \in \mathbb{R}^{L \times 1}$ for each path $m \in \{\mathtt{V}, \mathtt{T}, \mathtt{P}, \mathtt{E}\}$ in Figure~\ref{fig:visualization}. For example, in the video from the simulated dataset, frames with number \textit{1} record a miss of right robot hand while the surgeon is tightening suture, which leads to a simultaneous fall of event score. This error causes event repetition and interrupts the normal workflow of suturing. On the other hand, in the video from our clinical dataset, the frame with number \textit{2} presents a detected bleeding event and a concurrent fall of event score. In addition, the frame with number \textit{3} shows a camera out event with low tool scores. The camera out event likely corrupts the surgical tool segmentation algorithm, which can be regarded as a failure case.
A video demo is attached in the supplementary.

\section{How the Model Understands Surgical Skills}

To investigate what the model learns about surgical skills, the model outputs are analyzed quantitatively.
Note that some input feature sequences are of medical or physical meanings, such as the $X_{\mathtt{E}}$ on the simulated dataset indicating the probabilities of surgical gestures.
Therefore, we can get some insights by temporally correlating the weighted score sequences $S_{m}W_{m} \in \mathbb{R}^{L \times 1}$ to these meaningful input feature sequences.
If take $X_{\mathtt{E}}$ on the simulated dataset as an example, the correlation between model outputs and a channel $c$ in $X_{\mathtt{E}}$ is defined as $R_{\mathtt{E}}^{(c)}$ by the following equation:
\begin{equation}
R_{\mathtt{E}}^{(c)} = \frac{1}{3}\sum_{m} \lvert \mathrm{srocc}(S_{m}W_{m}, X_{\mathtt{E}}^{(c)}) \rvert, m \in \{\mathtt{V}, \mathtt{T}, \mathtt{E}\}
\end{equation}
where $X_{\mathtt{E}}^{(c)} \in \mathbb{R}^{L \times 1}$ is the selected channel of the input.
SROCCs between this channel and weighted score sequences from different paths are computed and averaged.
The resultant $R_{\mathtt{E}}^{(c)} \in [0,1]$ indicates to what extent the skill predictions are correlated to the surgical gesture $c$.
In the equation above, we take the absolute value of the SROCC since we care about the correlation regardless of whether it is positive or negative.
The $S_{\mathtt{P}}W_{\mathtt{P}}$ is excluded from the computation since it is a constant sequence.
Apart from $X_{\mathtt{E}}$, the tool feature sequence $X_{\mathtt{T}}$ also has physical meanings on the simulated data and $R_{\mathtt{T}}$ can be similarly computed.

\begin{figure}[t]
\begin{center}
   \includegraphics[width=1.0\linewidth]{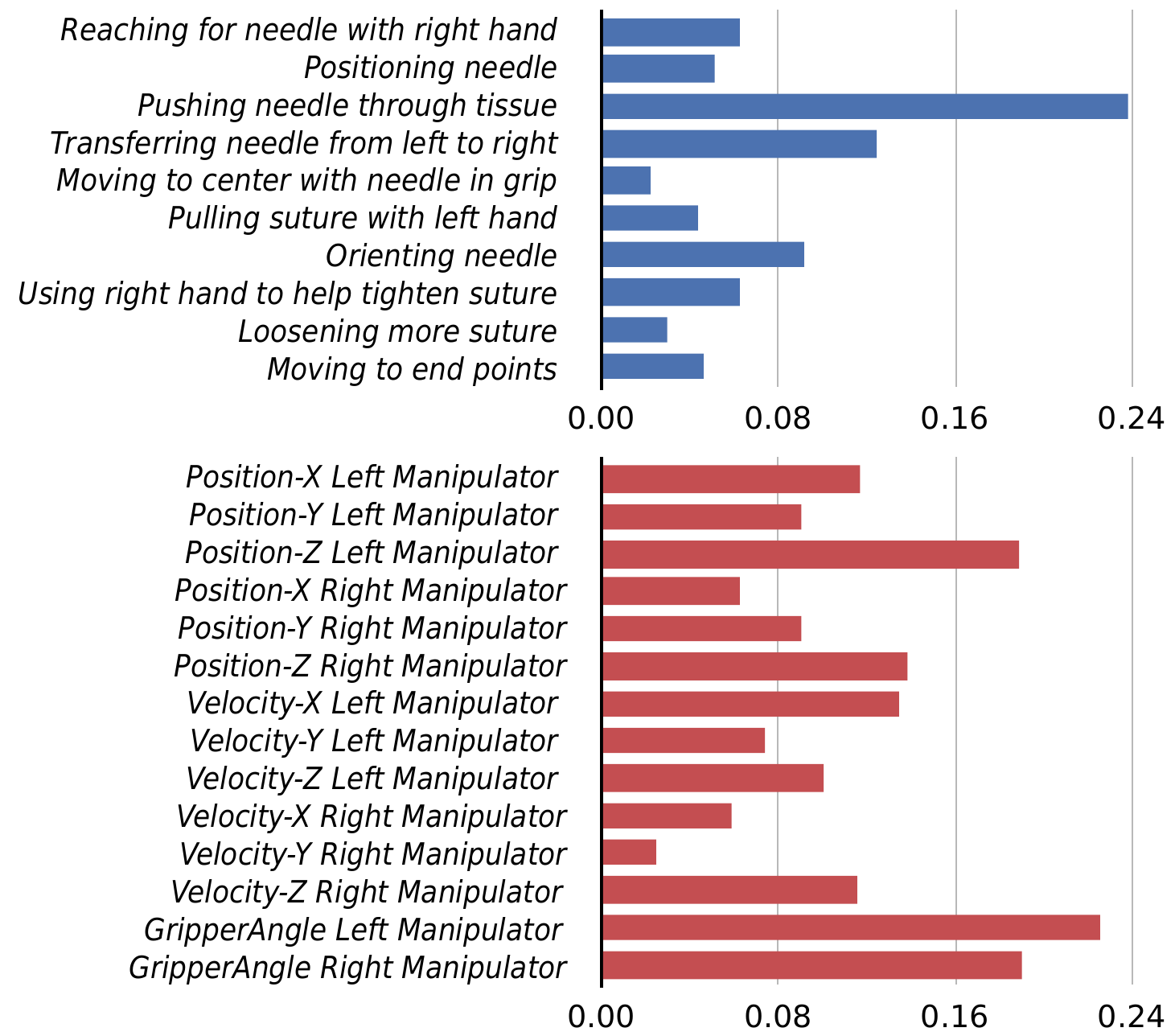}
\end{center}
   \caption{Blue: Correlations between model outputs and surgical gestures on the simulated suturing ($R_{\mathtt{E}}$). Red: Correlations between model outputs and tool features on the simulated suturing ($R_{\mathtt{T}}$).}
\label{fig:insight_simulated}
\end{figure}

The bar plots of $R_{\mathtt{E}}$ and $R_{\mathtt{T}}$ for the simulated suturing task are given in Fig.~\ref{fig:insight_simulated}.
For $R_{\mathtt{E}}$, it is interesting that the model outputs are most correlated to the surgical gesture \textit{``Pushing needle through tissue"}, which is also an intuitively critical gesture for suturing.
Model outputs are less correlated to transitional gestures such as \textit{``Moving to center with needle in grip"} and \textit{``Moving to end points"}.
For $R_{\mathtt{T}}$, it is noticed that the model outputs are most correlated to the gripper angles and the position-z of the left manipulator, which are also highly active factors during suturing in practice.
These findings are consistent with the human understanding of suturing.

\begin{figure}[t]
\begin{center}
   \includegraphics[width=1.0\linewidth]{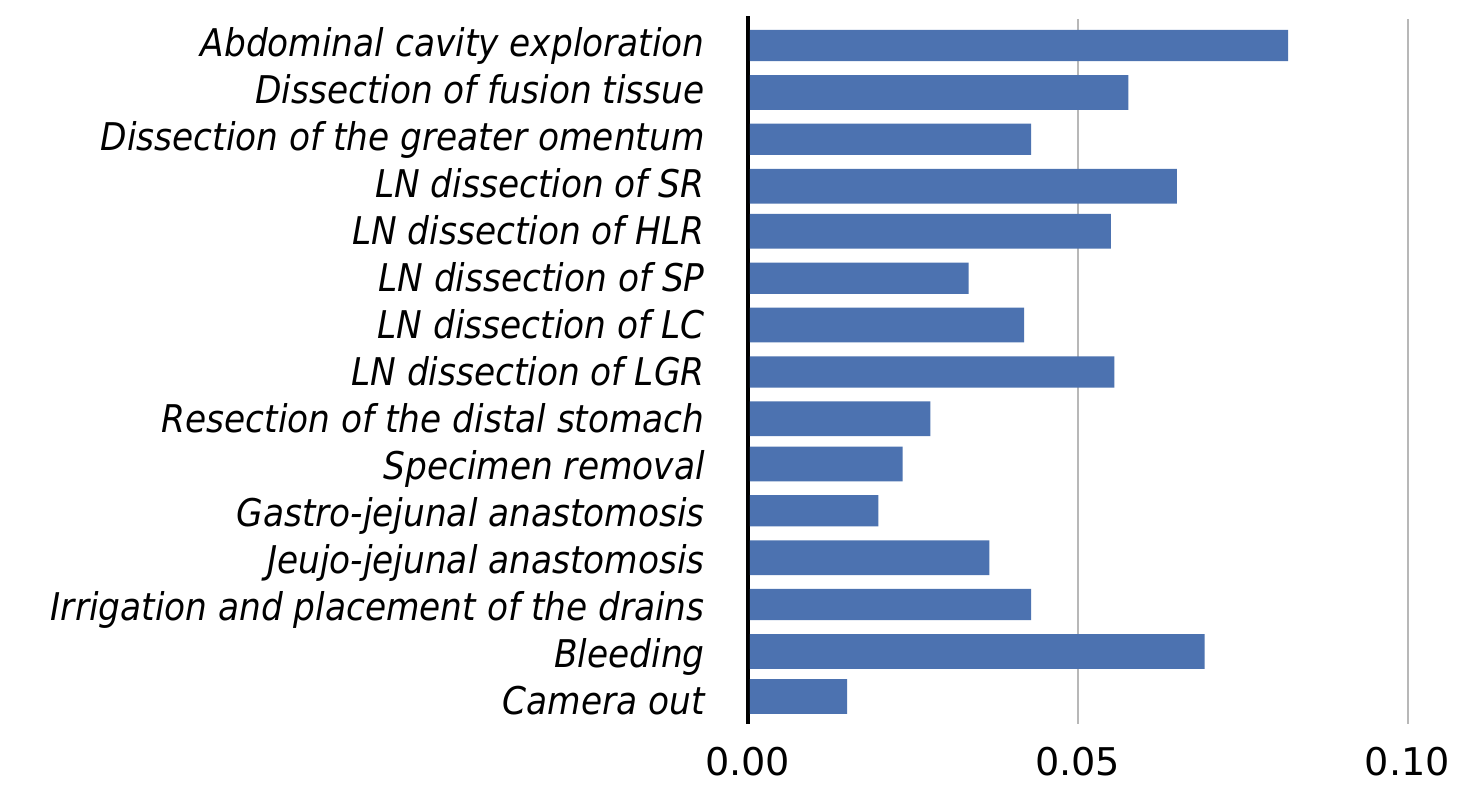}
\end{center}
   \caption{Correlations between model outputs and surgical events on the clinical data ($R_{\mathtt{E}}$).}
\label{fig:insight_clinical}
\end{figure}

Similarly, the $R_{\mathtt{E}}$ on the clinical data is also computed and visualized in Fig.~\ref{fig:insight_clinical}.
The $S_{\mathtt{P}}W_{\mathtt{P}}$ is included in the computation now.
Currently, no remarkable correlation between the surgical event and the model output is observed, with all correlations are lower than $0.1$.
This demonstrates the simulated-clinical gap and thus the importance of clinical data.
A larger number of clinical surgeries may lead to more evident findings in the future.
More results are in the supplementary.

\section{Conclusion and Future Work}
This paper proposes a flexible and general framework to automatically assess surgical skills from multiple aspects.
The effectiveness of the proposed framework is validated by the experiments on both simulated and clinical surgery datasets.
Within this framework, future works could focus on more advanced input features and composing functions.
Our framework is also extendable to include more skill aspects beyond those used in this study.
Future works could also research the flexible choice of paths and the adaptive fusion of paths.
Besides, it is also desirable to have more data from the clinical environment for surgical skill assessment in the future.

\textbf{Acknowledgments.} This work was partially supported by NSFC-61625201 and NSFC-62061136001. We also acknowledge the Clinical Medicine Plus X-Young Scholars Project, and High-Performance Computing Platform of Peking University for providing computational resources.

{\small
\bibliographystyle{ieee_fullname}
\bibliography{mybib}
}

\end{document}